\documentclass[lettersize,journal]{IEEEtran}
\usepackage{amsmath,amsfonts,bm}

\usepackage{algorithm,algorithmicx,algpseudocode}
\usepackage{array}
\usepackage[caption=false,font=normalsize,labelfont=sf,textfont=sf]{subfig}
\usepackage{textcomp}
\usepackage{stfloats}
\usepackage{url}
\usepackage{verbatim}
\usepackage{graphicx}
\usepackage{multirow,threeparttable}
\usepackage{epstopdf}
\usepackage{amsthm}
\usepackage{color,hyperref}

\usepackage[switch]{lineno}
\hypersetup{colorlinks = true, %将链接文字带颜色
	    linkcolor = black, %图表引用颜色设置为蓝色
	    urlcolor = blue,%网页链接为蓝色
           citecolor = blue} %文献引用颜色设置为蓝色

\def\BibTeX{{\rm B\kern-.05em{\sc i\kern-.025em b}\kern-.08em
    T\kern-.1667em\lower.7ex\hbox{E}\kern-.125emX}}
\usepackage{balance}
\begin{document}
%\linenumbers
\title{Rectify and Align GPS Points to Parking Spots via Rank-1 Constraint}

\author{Jiaxin Deng, Junbiao Pang, Zhicheng Wang, and Haitao Yu % <-this % stops a space
\thanks{
\IEEEcompsocthanksitem J. Deng, J. Pang and Z. Wang are with the Faculty of Information Technology, Beijing University of Technology, Beijing 100124, China (e-mail: \mbox{junbiao\_pang@bjut.edu.cn}).

\IEEEcompsocthanksitem   H. Yu is with the Beijing Intelligent Transportation Development Center,
Beijing 100161, China (email: yuhaitao@jtw.beijing.gov.cn).
}% <-this % stops a space
}

%\markboth{Journal of \LaTeX\ Class Files,~Vol.~18, No.~9, September~2020}%

\maketitle

\maketitle

\begin{abstract}

Parking spots are essential components, providing vital mobile resources for residents in a city. Accurate Global Positioning  System (GPS) points of parking spots are the core data for subsequent applications, \textit{e.g.}, parking management, parking policy, and urban development. However, high-rise buildings tend to cause GPS points to drift from the actual locations of parking spots; besides, the standard lower-cost GPS equipment itself has a certain location error. Therefore, it is a non-trivial task to correct a few wrong GPS points from a large number of parking spots in an unsupervised approach. In this paper, motivated by the physical constraints of parking spots (\textit{i.e.}, parking spots are parallel to the sides of roads), we propose an unsupervised low-rank method to effectively rectify errors in GPS points and further align them to the parking spots in a unified framework. The proposed unconventional rectification and alignment method is simple and yet effective for any type of GPS point errors. Extensive experiments demonstrate the superiority of the proposed method to solve a practical problem. The data set and the code are publicly accessible at:~\url{https://github.com/pangjunbiao/ITS-Parking-spots-Dataset}.
\end{abstract}


% make the title area
\maketitle

\IEEEdisplaynotcompsoctitleabstractindextext
\IEEEpeerreviewmaketitle

\section{Introduction}\label{sec:intro}

\newtheorem{myobr}{Observation}
\newtheorem{mydef}{Definition}
\newtheorem{mythe}{Theorem}
\newtheorem{mypro}{Proposition}

Currently, parking services are integrated components of urban green infrastructure and significantly influence urban mobility ecosystem dynamics. Precise geo-spatial localization of parking facilities through GPS points and their subsequent integration into digital mapping platforms are essential for multi-scale viewing and analyzing for different applications, \textit{e.g.}, parking management, navigation services, and data-driven decision-making. Therefore, it is
desirable to develop an efficient, and cost effective way to create precise GPS locations or improve
existing ones~\cite{l2mm}.

{\color{blue} In urban areas, however, GPS points often mismatch the actual locations of Points Of Interest (POIs), due to obstructions from high-rise buildings, the error tolerance of GPS equipment~\cite{2021MAP}, various electronic interferences, multi-path errors~\cite{Liu2023Discrete} and Non-Line-Of-Sight (NLOS) signal~\cite{2023NLOS}, or even mis-operation of GPS collectors~\cite{2022Guide}. Consequently, some GPS points of parking spots mismatch the true positions. For example, commonly used single-frequency GPS receivers have an accuracy of around 7.8 meters, while more expensive GPS dual-frequency receivers have provided an accuracy of approximately 0.715 meters~\cite{Fahim2021Smart}.
While the majority of GPS coordinates exhibit high positional accuracy, sporadic mismatches significantly impair the downstream functionality of Intelligent Transportation System (ITS), \textit{e.g.}, digital transportation~\cite{tan-trip-2024-digital-transportation} and parking planning~\cite{2023AElfaki}.
To address this challenge, we introduce an unsupervised, yet universal methodology for GPS point rectification and alignment specifically tailored to parking spots. Our approach enhances the geo-spatial precision of parking spot identification in complex urban environments, thereby improving the reliability of downstream ITS applications.}

%这个引用加进第一段
%However, the positioning effect of smartphones in urban environments is not ideal, and the non-line-of-sight (NLOS) signal is one of the main limiting factors.~\cite{2023NLOS}~\cite{Liu2023Discrete}

\begin{figure}[t!]
\centering
\subfloat[Translational error]
{\includegraphics[width=6.5cm, height=1.7cm]{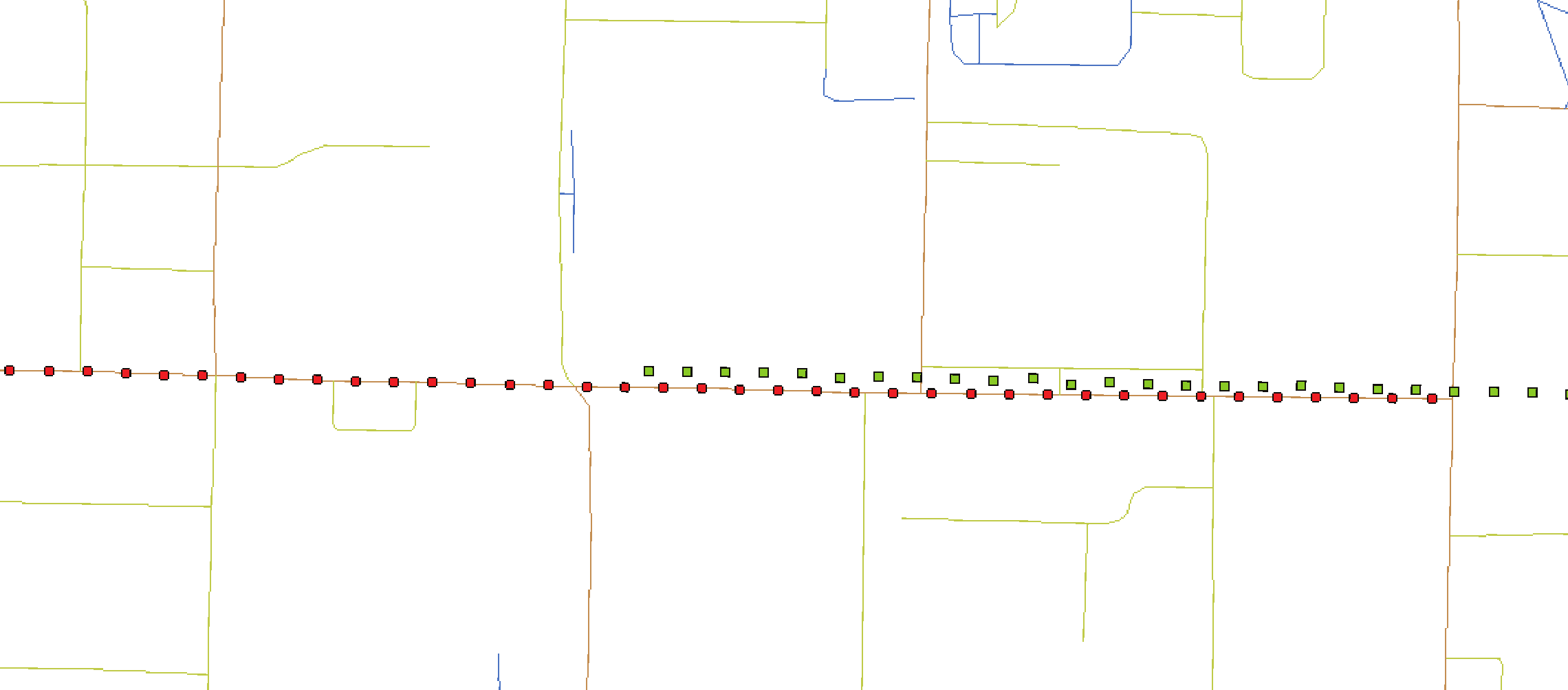}
}\\
\subfloat[Rotational error]
{\includegraphics[width=6.5cm, height=1.7cm]{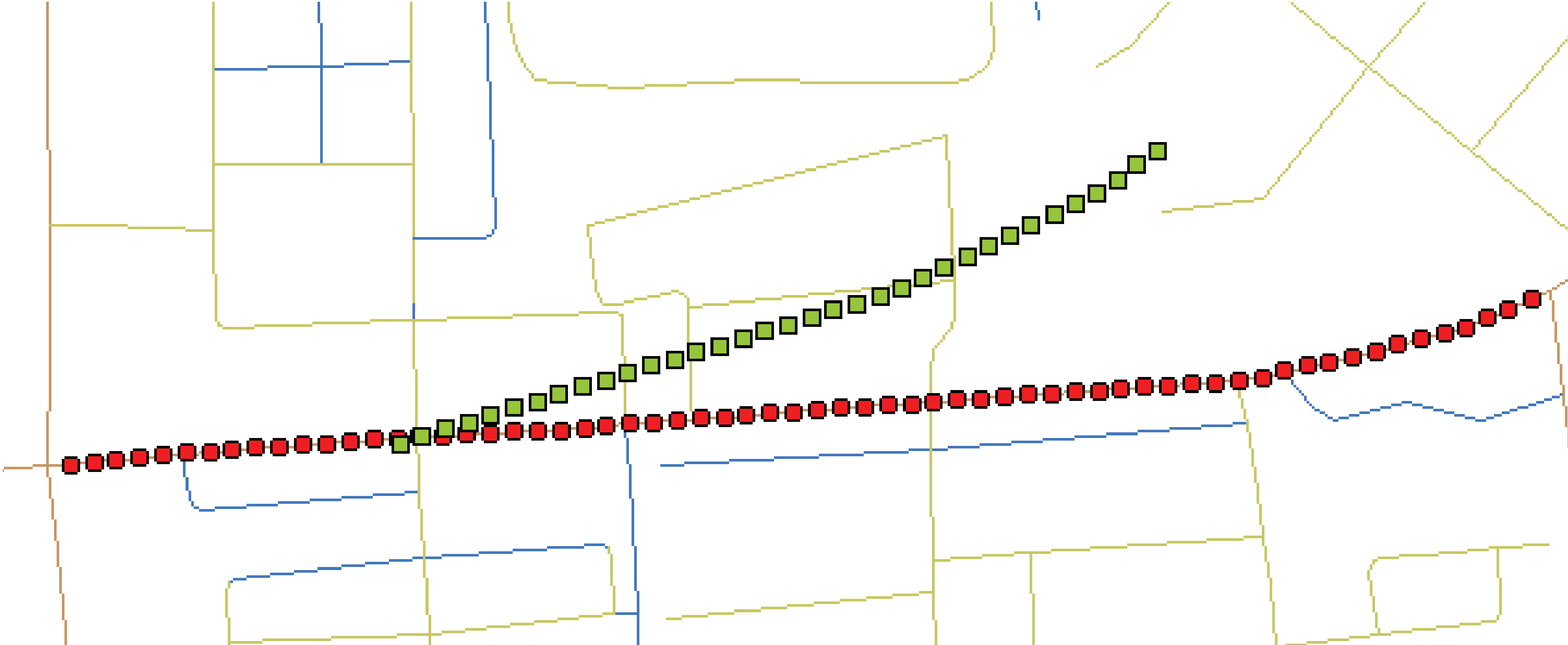}
}\\
\subfloat[Random error]
{\includegraphics[width=6.5cm, height=1.7cm]{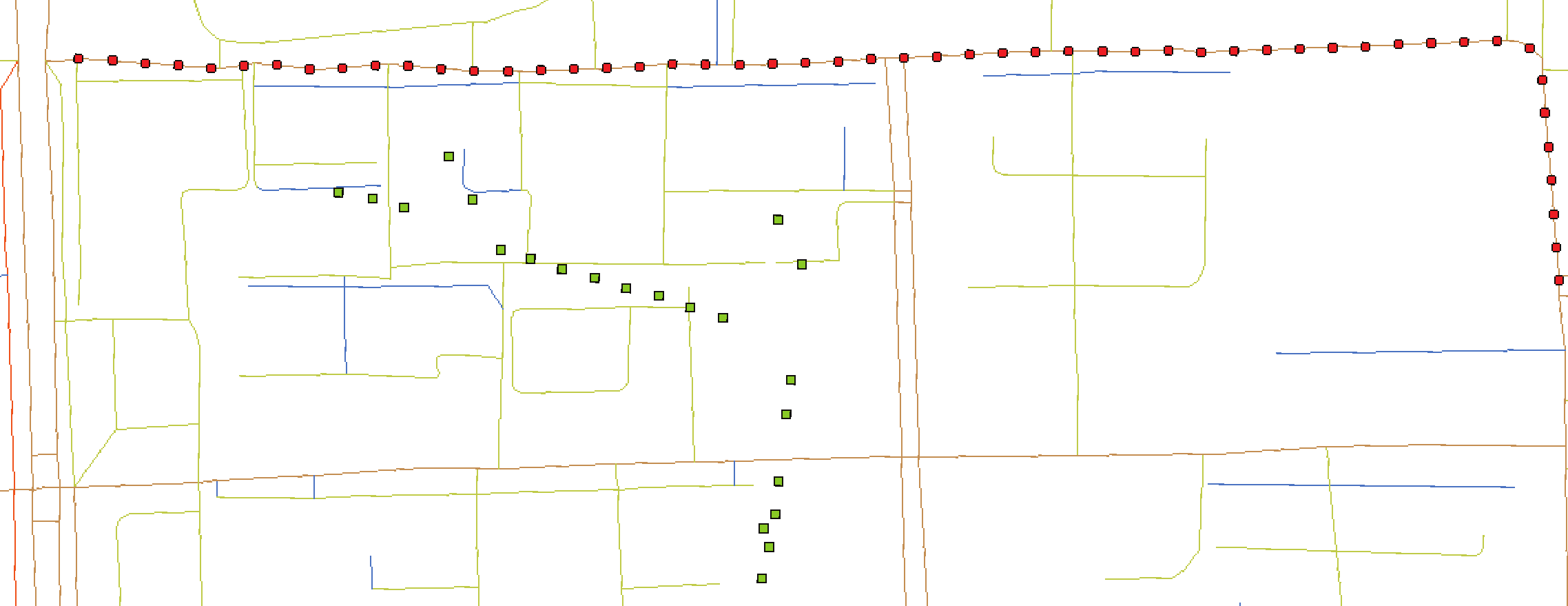}
}\\
\subfloat
{\includegraphics[width=6cm, height=0.25cm]{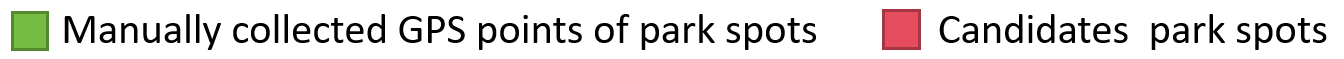}
}
\caption{Representative examples of corrupted GPS points and the corresponding parking spots (best viewed in color).}\label{fig:problem}
\end{figure}

%{\color{blue} In practice, we only have the GPS information of streets in a city rather than the ones of parking spots. To measure whether manually collected GPS points are correct ones for parking spots or not, we assume that legitimate parking spots must adhere to the geometric constraints imposed by street infrastructure. Specifically, we employ interpolation on GPS data of a road segment to generate candidate parking positions. Any manually collected GPS point that fails to align with this candidate set is flagged as a potentially error GPS points. }

{\color{blue} Our analysis of the manually curated dataset in Section~\ref{sec:Experimental} reveals the following important observations: 1) the majority of GPS points exhibit precise alignment with reference road segments; and 2) only a few GPS points show statistically significant positional deviations from real locations. Additionally, we categorized three predominant error types observed in the dataset:
  \begin{itemize}
\item  \textbf{Translational error.} The spatial coordinates of parking locations derived from the road segment exhibit systematic translational displacement relative to the road segment centerline, as shown in Fig.~\ref{fig:problem}(a).
\item  \textbf{Rotational error.} The orientation of collected GPS points of parking spots exhibits a significant rotational bias relative to the road segment alignment, as shown in Fig.~\ref{fig:problem}(b).
\item  \textbf{Mixed error.} The positional inaccuracies of GPS points arise from a composite of translational, rotational, and stochastic errors. Notably, the stochastic error manifests as substantial deviations or omissions in GPS coordinates relative to the true parking locations, as shown in Fig.~\ref{fig:problem}(c).
\end{itemize}
}

Three key challenges can be derived from Fig.~\ref{fig:problem} as follows: First, the mixture of three types of GPS errors complicates the correction of erroneous points by a supervised approach, as establishing point-wise correspondences for supervised learning is difficult. Second, although some GPS points are lost or drifted under mixed errors, our objective is to not only reuse but also rectify these erroneous GPS points. {\color{blue} ‌Third, it is imperative to implement a comprehensive error-handling mechanism capable of addressing multiple error types concurrently, given their potential to manifest independently or in synergistic combinations. } The indistinguishability among these error types constitutes a fundamental challenge in developing an efficient rectification framework for erroneous GPS points, particularly in achieving precise alignment with Points of Interest (POIs)~\cite{bougie2024lost}.

%xxxx为什么不能用supervised approach？1.大多数都是都是匹配的，只有部分不匹配。2.对于不匹配的，手工收集点和road chain sampling难以建立单个的点的对应关系形成有监督所需的标签。

{ \color{blue} Conventionally, mapping GPS points to the true POIs is seemingly similar to the classical map-matching problem~\cite{millard2019map} in different scenarios, \textit{e.g.}, urban traffic modeling~\cite{castro2012urban}~\cite{ yu2011T}, mobility pattern mining~\cite{kim2015TOPTRAC}~\cite{jeung2011Trajectory} and dynamic road map generation~\cite{liu2012mining}. However, the traditional map-matching methods, \textit{i.e.}, Chamfer Distance (CD)~\cite{veeravasarapu2020proalignnet} matching and Euclidean distance (ED) matching, do not effectively address the challenges of correcting erroneous GPS points for parking spots. Because these traditional map-matching methods focus on aligning the tracks of a vehicle to a reasonable road segments for the visual displacement purpose, overlooking whether the mapped points are precise or not. In contrast, correcting GPS points for parking spots focuses on accurately positioning a GPS point within a specific parking spot, rather than just aligning them to a road segment.

% 我们想要什么样的算法, 按照这个思路改写
%this paper, we propose an universal unsupervised method to the Rectifying-And-Aligning (RAA) GPS points for parking spots.
%xxxx我们需要解决3种问题的无监督方法-》但是现存方法无法解决这三种问题，为什么-》因此我们提出。。。解决Rectifying-And-Aligning (RAA) problem-》下一段具体

Therefore, it is natural to ask whether GPS points of parking spots could be accurately rectified for all types of errors in Fig.~\ref{fig:problem}, and even beyond, by an unsupervised approach. Unsupervised methods offer several key advantages: good generalization ability, elimination of time-consuming manual annotation, and a unified model for handling all type errors.}
In this paper, we propose a low-rank method to effectively Rectify And Align (RAA) manually collected GPS points for parking spots, assuming that the matrix built from both the parking spots and the GPS points should be rank 1. When  GPS points are correctly mapped to their corresponding parking spots, the rank of the matrix formed by these parking spots and the GPS points will naturally reduce to 1. To preserve the geometric relationship between the parking spots and the GPS points, we incorporate a noise rectification component and a rotation-translation operation, which together correct and align the GPS points to their corresponding parking spots. Experimental results demonstrate that our method effectively handles the error types in Fig.~\ref{fig:problem} in a unified approach.

To the best of our knowledge, this is the first method to rectify and align GPS points and parking spots using an unsupervised approach. The proposed method is universal and highly effective in rectifying and aligning two point sets in an unsupervised approach. By assuming low-rank, \textit{i.e.}, rank-1 constraint, without distinguishing the type of errors in Fig.~\ref{fig:problem}, we present a novel method that matches and even outperforms the traditional approach~\cite{borgefors1988hierarchical}~\cite{veeravasarapu2020proalignnet}~\cite{danielsson1980euclidean} on the public dataset.

\section{related work}\label{sec:relatedwork}
\subsection{Map-matching}
%细分方法
%from these paper A Survey on Map-Matching Algorithms ,and L2MM: Learning to Map Matching with Deep Models for Low-Quality GPS Trajectory Data ~\cite{l2mm},
Map-matching~\cite{ASurvey2020} projects the collected GPS points onto a road segment for identifying the location of a vehicle on a road network. The map-matching algorithms are the key component to improve the performance of systems that support the navigation of ITS~\cite{quddus2007current}.
Influenced by challenging environments~\cite{2017Accurate}, GPS trajectories often suffer from data quality issues, \textit{e.g.}, noise, low-frequency, and tolerance of a GPS equipment itself. Consequently, map-matching focuses on handling with noisy, and low-frequency GPS data by either conventional model-based methods or emerging learning-based models.

The model-based methods use various information, \textit{e.g.}, spatial-temporal features~\cite{2017AST}, heading direction~\cite{2019TrajCompressor}, moving speed~\cite{2016IF}, trajectory similarity~\cite{2020ATra}, and driver behavior~\cite{2018Feature}, to find reasonable mapping . For instance, Hidden Markov Model (HMM) and its variants~\cite{2019TrajCompressor} are widely used for mapping low-frequency data. However, the model-based method belongs to the supervised method, which is barely readily extended to RAA parking spots~\cite{2013MM}.

{\color{blue}
Learning-based models have been developed in a data-driven fashion. For example, DeepMM~\cite{2020DeepMM} was proposed to match sparse and noisy trajectories. More recent works, such as \cite{jin2022transformer}~\cite{mohammadi2024Trajectory}, have employed Transformer-based encoder-decoder method, treating map-matching as an NLP-like task to enhance performance in complex urban road networks. RLOMM \cite{chen2025rlomm} was developed for efficient and robust online map matching.
However, these learning-based methods are typically supervised approaches that are not only the task-specific approach but also rely on large amounts of labeled data. For instance, transformer-based model~\cite{mohammadi2024Trajectory} heavily relied on some tailed component (\textit{e.g.}, trajectory generation, and fine-tuning technique) and used 35,127 GPS points. In contrast, our method is an unsupervised approach which efficiently leverages a small GPS points, but is robust to different GPS errors.
}

\subsection{Low-Rank Assumption}

Low-rank implicitly assumes that the system is captured by a small number of variables, which are sufficient to describe the underlying network of interactions~\cite{pang-its-matrix-decom-18}~\cite{2024low-rank}. For instance, Robust Principal Component Analysis (RPCA)~\cite{2002Robust}~\cite{2010TILT} decomposed a matrix into a low-rank matrix and a sparse matrix, achieving the purpose of noise removal and obtaining the low-dimensional components.

{\color{blue} In this study, we employ the rank-1 constraint to establish correspondence between collected GPS points and parking spots. This constraint offers a distinct advantage over point-wise comparison methods (\textit{e.g.}, Euclidean distance minimization) by preserving the geometric ``shape'' grouped from parking spots, rather than merely achieving averaged distance minimization between two point sets. To our knowledge, this is the first to introduce a low-rank constraints for both rectification and alignment of parking spot GPS points.}

\section{OUR APPROACH}\label{sec:approach}

\subsection{Background of Parking Spots}\label{subsec:backgroundofapproach}

\begin{figure}[t!]
    \centering
    \includegraphics[width=0.9\linewidth]{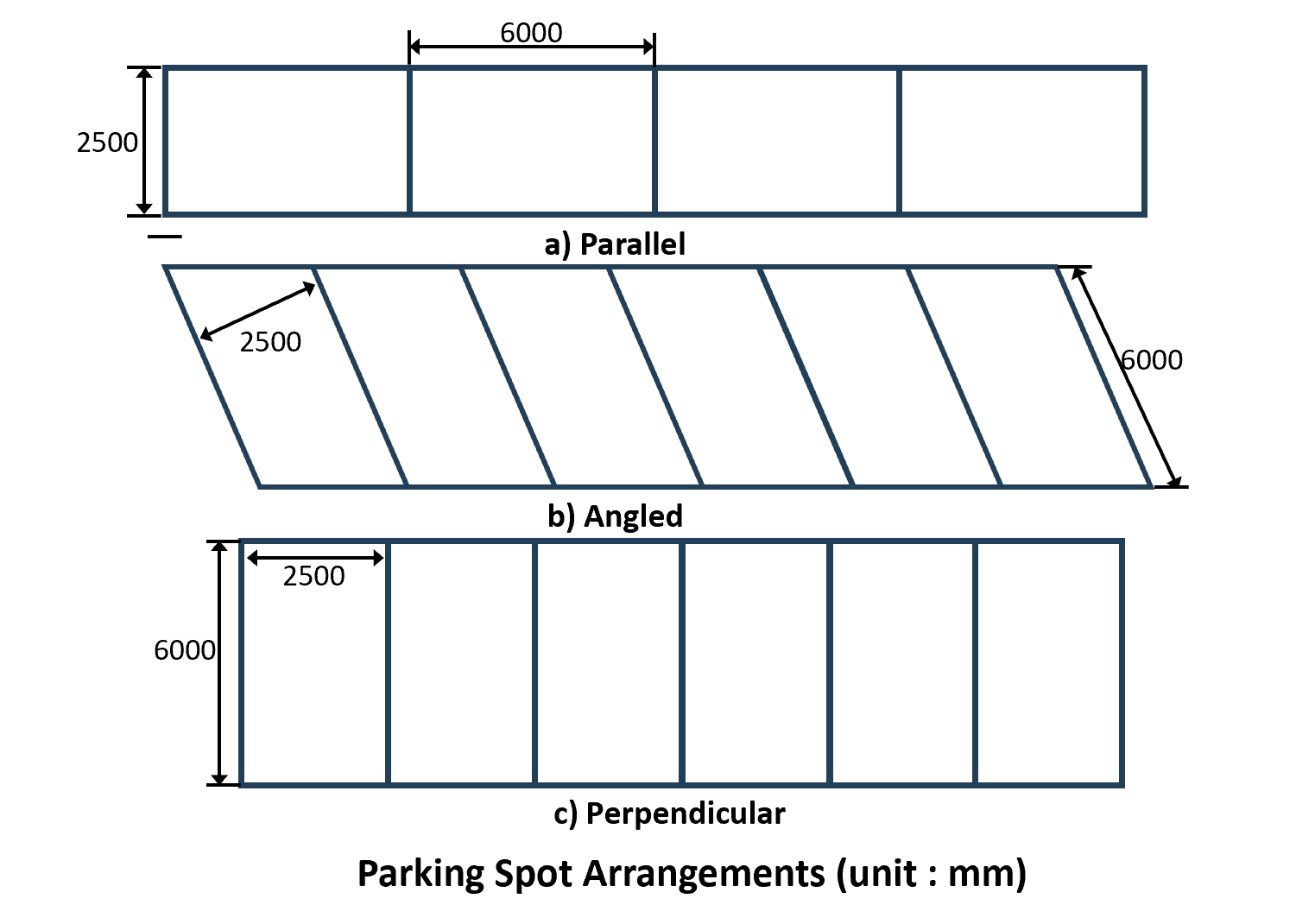}
    \caption{The layout of three types parking spots.}
    \label{fig:parking-spot-arrangements}
\end{figure}

\textbf{Physical constraints of parking spots.} {\color{blue}%Parking spots have a rectangle shape with a width of 2.5 meters and a length of 6 meters.
Parking spots are typically arranged in three layouts, \textit{i.e.}, parallel, angled (with tilt angles of $30^\circ$, $45^\circ$, and $60^\circ$), and perpendicular, as illustrated in Fig.~\ref{fig:parking-spot-arrangements}. These configurations follow the standards for parking layouts in China.}
Specifically, the distance between two adjacent parking spots arranged in the parallel layout is 6 meters apart with 2.5 meters width at the sides of a road. The distance between the two inclined (or perpendicular) parking spots is 3 meters. Besides, according to parking spot standards, no parking spots are allowed within 50 meters of an intersection. {\color{blue} The road segment is assumed to represent the street centerline, reflecting the common real-world scenario where parking spots are symmetrically arranged on one or both sides of the road.} %In summary, a rectification method should follow the above two physical constraints.

\begin{mydef}[A road segment]\label{def:roadnework}
 A road segment in a map is a sequence of ordered GPS points, denoted as $\mathcal{S}=\{\mathbf{p}_1,\ldots, \mathbf{p}_i,\ldots, \mathbf{p}_N\}$, $1\leq i \leq N$, where a GPS point is $\mathbf{p}_i=(lat_i, lon_i)$, in which $lat_i$ and $lon_i$ correspond to the latitude and longitude, respectively. Besides, the sampling interval is defined as the spatial distance $\Delta d$ ($\Delta d >0$) between two consecutive GPS points $p_i$ and $p_{i+1}$.
\end{mydef}

%xxxx为什么要candidate %The sampled GPS candidate points pave the way to the proposed unsupervised method.
\textbf{Sampling GPS candidates of parking spots on a map.} We sample all possible GPS candidate points of parking spots by leveraging the physical constraints of parking spots.
{\color{blue}Specially, the interval $\Delta d$ in a road segment in Def.~\ref{def:roadnework} is typically longer than the length of a parking spot. We used linear interpolation to generate possible GPS candidates between GPS points $\mathbf{p}_i$ and $\mathbf{p}_{i+1}$ on a road segment, and subsequently filter out the interpolated GPS points that violate the following two physical constraints as follows:
\begin{itemize}
    \item \textbf{The type of parking spots.} We interpolated the intervals according to the type of parking spots. Concretely, 6 meters is used for the type 1, and 3 meters for the type 2 (type 3) parking spots.
    \item \textbf{50 meters regularization.} We remove the GPS points that violate the 50 meters regularization at intersections.
\end{itemize}
The possible GPS candidate points are denoted as $\mathcal{C}=\{\mathbf{p}_1,\ldots, \mathbf{p}_K\}$. Note that the number of sampled GPS candidates $\mathcal{C}$ is typically much larger than that of parking spots.} %An example of sampling candidate parking spots on one of the roads is shown in Fig.~\ref{fig:sampled-GPS}.

\subsection{Problem Definition}

\begin{mydef}[The RAA of GPS points for parking spots]\label{def:problem}
The objective of the RAA method for GPS points in parking spots is to find a function $f(\cdot)$ that rectifies error GPS points from the manually collected GPS points, $\mathcal{R}=\{\mathbf{p}_1,\ldots, \mathbf{p}_M\}$, corresponding to $M$ parking spots, as follows:
\begin{equation}\label{eqt:problem}
\begin{split}
\min_{\mathcal{\hat{R}}} \  \ &\text{dist}(\mathcal{\hat{R},G})\\
s.t.: \  \ & f\left(\mathcal{R ,C}\right) \rightarrow \mathcal{\hat{R}},
\end{split}
\end{equation}
where $\mathcal{C}=\{\mathbf{p}_1,\ldots, \mathbf{p}_K\}$ is the set of sampled GPS candidates for parking spots where $K \ge M$, and $\mathcal{G}= \{\mathbf{p}_1,\ldots, \mathbf{p}_M\}$ is the set of  ground truth GPS points of parking spots, (\textit{i.e.}, $\mathcal{G} \in \mathcal{C}$), $\hat{\mathcal{R}}=\{\hat{\mathbf{p}}_1,\ldots, \hat{\mathbf{p}}_M\}$ represents the rectified GPS points for the parking spots, and the function $dist(\mathcal{A,B})$ calculates the distance between the sets $\mathcal{A}$ and $\mathcal{B}$. Note that the ground truth set $\mathcal{G}$ is unavailable in practice.
\end{mydef}

%\begin{figure}
%    \centering
%    \includegraphics[width=0.8\linewidth]{figures/macth flow chart.jpg}
 %   \caption{{\color{blue} An overview of the flowchart for the RAA.}}
 %   \label{fig:enter-label}
%\end{figure}

{\color{blue}
\begin{small}
\begin{algorithm}[!t]
    \renewcommand{\algorithmicrequire}{\textbf{Input:}}
    \renewcommand{\algorithmicensure}{\textbf{Output:}}
    \caption{The unsupervised RAA for parking spots}
    \label{alg:1}
    \begin{algorithmic}[1]
        \Require
        The collected GPS points $\mathcal{R}=\{\mathbf{p}_1,\ldots, \mathbf{p}_M\}$; % input
        the road id $Id_R$; the error threshold $th$;
        GPS candidate points $\mathcal{C}=\{\mathbf{p}_1,\ldots, \mathbf{p}_K\}$ by the method in Section \ref{subsec:backgroundofapproach};
        \State Compute Euclidean distance between $\mathcal{R}$ and the first $M$ GPS points in  $ \bar{\mathcal{C}}\leftarrow \{\mathbf{p}_1,\ldots, \mathbf{p}_{M}\}\in \mathcal{C}$, \textit{i.e.}, $d= dist (\mathcal{R}, \bar{\mathcal{C})}$;
        \If {$d< th$} % good match
         \State $\mathcal{R} $ are correct GPS points;
        \Else % our method
        \For{$i = 1$ to $K-M$} %更换起点进行匹配
           \State Select the subset $\bar{\mathcal{C}}_i=\{\mathbf{p}_i,\mathbf{p}_{i+1},\ldots, \mathbf{p}_{i+M}\}$ as the candidate GPS points;
           \State Use Alg.~\ref{algorithmADMM} to align $\mathcal{{R}}$ and $\mathcal{\bar{C}}_i$ and record the alignment loss, \textit{i.e.}, $\|\bm{E}_1\|_1+\|\bm{E}_2\|_1+ \|\bm \Theta_1\|_2$;
        \EndFor
        \EndIf
    \Ensure The matched set $\mathcal{\bar{C}}_i$ with the lowest alignment loss. % output
 \end{algorithmic}
\end{algorithm}
\end{small}

The proposed RAA approach is summarized in Alg.~\ref{alg:1}. First, the manually collected GPS points for the parking spots $\mathcal{R}$ and the corresponding road segments are retrieved. Second, GPS candidate points are sampled from the road segments. Third, we check whether the mismatch between the collected points and the candidates is larger than a predefined threshold or not. Fourth, if a mismatch is detected, we adopt a brute force approach to find the best GPS parking points; concretely, we extract a subset of GPS points $\mathcal{\bar{C}}=\{\mathbf{p}_i,\ldots, \mathbf{p}_{i+M}\} (i+M \leq K)$ with the same size as $\mathcal{R}$, and apply the rank-1 alignment method.
}

\subsection{Rank-1 Alignment}

We assemble both the latitudes and longitudes of the manually collected GPS points set $\mathcal{R}=\{\mathbf{p}_1,\ldots, \mathbf{p}_M\}$ into the matrix $\bm{P} \in \mathbb{R}^{2M \times 1}$. Similarly, the subset of candidate parking spots $\mathcal{\bar{C}}=\{\mathbf{p}_i,\ldots, \mathbf{p}_{i+M}\}$ is converted into the matrix $\bm{R}_d \in \mathbb{R}^{2M \times 1}$. To deal with the errors in GPS points, we introduce the sparse error matrix $\bm {E}_1 $ ($\bm {E}_1 \in \mathbb{R}^{2M \times 1}$), the translation error matrix $\bm{E}_2$ ($\bm {E}_2 \in \mathbb{R}^{2M \times 1}$), the 2-Dimensional (2D) homogeneous matrix $\bm{\Theta}_1$  ($\bm {\Theta}_1 \in \mathbb{R}^{3 \times 3}$) and $\bm{\Theta}_2$ ($\bm {\Theta}_2 \in \mathbb{R}^{3 \times 3}$) to rectify the error GPS points as follows:
\begin{equation}\label{eqt:rank-1-align}
\begin{split}
\min_{\bm {E}_1,\bm{E}_2,\bm{\Theta}_1,\bm{\Theta}_2} &  \| \bm {E}_1\|_1+\|\bm{E}_2\|_2 + \lambda\|\bm A\|_{rank=1} \\
\text{s.t.:} \quad&  \bm{\Theta}_1 \circ  \bm P+\bm E_1  =\bm C \\
      &    \bm{\Theta}_2  \circ \bm{R}_d+\bm E_2 = \bm D \\
      &
\begin{bmatrix}
\  \bm C  \\
   \bm D
\end{bmatrix} = \bm A,
\end{split}
\end{equation}
where $\bm{\Theta}_1 \circ \bm P$ denotes GPS point warping, which applies the rotation and translation operation $\bm{\Theta}_1$ to all GPS points in $\bm P$, similarly for $\bm{\Theta}_2 \circ \bm{R}_d$. $\bm{C} \in \mathbb{R}^{2M \times 1}$ and $\bm{D} \in \mathbb{R}^{2M \times 1}$ are the rectified GPS points and GPS candidate points, respectively. $\lambda$ ($\lambda >0$) is a hyper-parameter. $ A \in \mathbb{R}^{2M \times 2} $ is the column concatenation of matrices $\bm{M}$ and $\bm{D}$, with an initial rank of 2.

\begin{figure}
   \centering
   \includegraphics[width=1\linewidth]{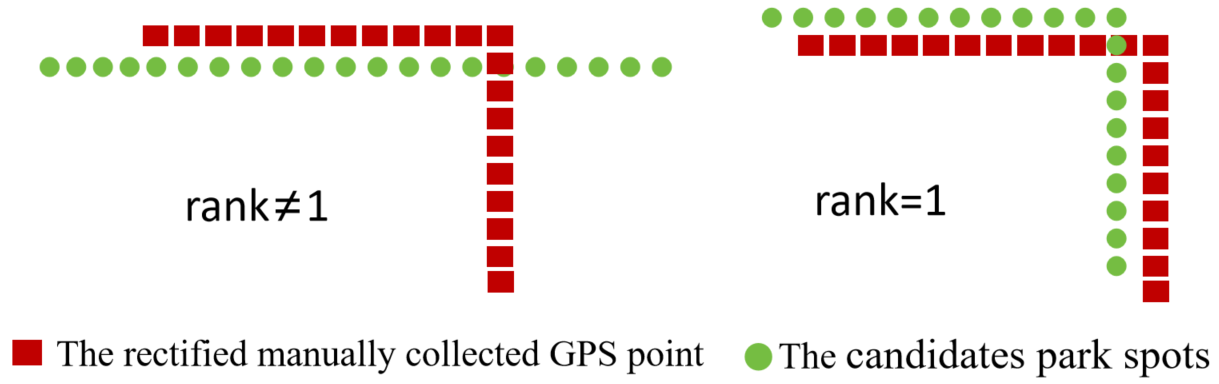}
   \caption{Rank-1 constraint enables GPS point alignment by preserving dominant spatial patterns while maintaining structural integrity.}\label{fig:rank1}
\end{figure}

{\color{blue} Despite some GPS points have errors, the structure of GPS points for parking spots along a road exhibits a predominantly linear or curvilinear pattern. This observation implies that the  matrix $\bm{A}$ is low-rank. When $\bm{A}$ has rank 1, it indicates that every column of $\bm{A}$ is a linear combination of a single pattern. Consequently, the rank-1 constraint aligns GPS points in both $\bm{C}$ and $\bm{D}$ along a dominant spatial pattern without losing the structure pattern as illustrated in Fig.~\ref{fig:rank1}. %Therefore, we enforce the rank-1 constraint on the matrix $\bm{A}$ to align the rectified GPS point matrix $ \bm C $ with the candidate matrix $ \bm D $, \textit{i.e.}, $ \|\bm{A}\|_{\text{rank}=1} $.

The sparsity constraint $\|\cdot\|_1$ in Eq.~\ref{eqt:rank-1-align} forces the matrix $\bm{E}_1$ to model the sparse errors in the GPS points matrix $\bm{P}$; while we empirically found that $\|\bm{E}_2\|_1$ on the matrix $\bm{R}_d$ promotes faster convergence of Alg.~\ref{algorithmADMM}. The 2D homogeneous matrix $ \bm{\Theta}_1 $ is as follows:
\begin{equation}
\bm{\Theta}_1 = \left[\begin{array}{ccc}
  cos\theta_1  & -sin\theta_1 &  s_{x_1}\\
    sin\theta_1 & cos\theta_1  & s_{y_1}\\
    0& 0& 1
\end{array}
\right],
\end{equation}
where $\theta_1$ is the rotation angle, $s_{x_1}$ and $s_{y_1}$ are the translation along the $x$ axis and $y$ axis, respectively. The operation $ \bm{\Theta}_1 \circ  \bm P $ is for all the point $\mathbf{p}_i$ in the matrix $\bm P$ as follows:
\begin{eqnarray}
    p_i &\leftarrow p_i cos\theta_1 -p_{i+1}sin\theta_1+s_{x_1}\\
    p_{i+1}&\leftarrow p_i sin\theta_1 +p_{i+1}cos\theta_1+s_{y_1},
\end{eqnarray}
where $p_i$ and $p_{i+1}$ are the latitudes and longitudes of $i$-th GPS point $\mathbf{p}_i=(p_i,p_{i+1})$, respectively. $\bm{\Theta}_2$ follows a similar definition as $\bm\Theta_1$, with $\theta_1$, $s_{x1}$ and $s_{y_1}$ replaced with $\theta_2$, $s_{x2}$ and $s_{y_2}$.
}

%\begin{figure}
%    \centering
%    \includegraphics[width=1\linewidth]{2-3.png}
%    \caption{Schematic diagram of MALR model matching process}\label{fig:2_3}
%\end{figure}

\subsection{Iterative Convex Optimization}\label{sec:optimization}

We assume that when the change in $\bm\Delta \bm\Theta_1$ is very small, the change from $\bm\Theta_1$ to $\bm\Theta_1 + \bm\Delta \bm\Theta_1$ can be linearized around the current estimation. Thus, the first constraint in Eq.~\eqref{eqt:rank-1-align} is simplified as follows:
\begin{equation}\label{eq:linearlized-roation}
\bm\Theta_1 \circ \bm{P}+ \bm\Delta\bm\Theta_1\circ\bm\nabla \bm{P} +\bm{E}_1=\bm{C},
\end{equation}
where $\bm\nabla \bm{P} $ is the Jacobian matrix of the GPS coordinates \textit{w.r.t}
the transformation parameters. Similarly, the operation $\bm\Theta_2 \circ \bm{R}_d$ is also linearized. Eq.~\eqref{eqt:rank-1-align} is converted as follows:
\begin{equation}\label{eq:linerlized-objective}
\begin{split}
\min_{\bm {E}_1,\bm{E}_2,\bm{\Theta}_1,\bm{\Theta}_2} & \ \ \| \bm E\|_1+\|\bm{E}\|_2 +\lambda\|\bm A\|_{rank=1} \\
\text{s.t.:} \quad
& \bm\Theta_1 \circ \bm{P} + \bm\Delta\bm\Theta_1 \circ \bm\nabla \bm{P} + \bm{E}_1 = \bm{C} \\
& \bm\Theta_2 \circ \bm{R}_d + \bm\Delta  \Theta_2\circ\bm\nabla\bm{R}_d + \bm{E}_2 = \bm{D} \\
&
\begin{bmatrix}
\bm{C} \\
\bm{D}
\end{bmatrix}
= \bm{A}.
\end{split}
\end{equation}

The linearized problem in Eq.~\eqref{eq:linerlized-objective} is convex and amenable to an efficient solution~\cite{grussler2018low}. Since the linearization is only a local approximation to the original nonlinear problem, we solve it iteratively in order to converge to
a (local) minimum of the original non-convex problem. The Alternating Direction Method of Multipliers (ADMM) is a classical algorithm that simultaneously minimizes the augmented Lagrangian function and computes appropriate Lagrange multipliers. We solve Eq.~\eqref{eq:linerlized-objective} by the ADMM as follows:

\begin{equation}\label{eq:lagarian-multiplier-objective}
\begin{split}
L & =\left\|\bm{E}_1\right\|_1 + \left\|\bm{E}_2\right\|_2 + \lambda\|\bm{A}\|_{rank=1} \\
& +\langle \bm{Y}_1, \bm\Theta_1    \circ\bm{P}+\bm{E}_1+\bm\Delta \bm{\Theta}_1\circ\bm \nabla \bm{P} - \bm{C}\rangle \\
& +\mu / 2\left\|\bm\Theta_1 \circ \bm{P}+\bm{E}_1+\bm\Delta \bm{\Theta}_1 \circ\bm\nabla \bm{P} -\bm{C}\right\|_F^2 \\
& +\langle \bm{Y}_2, \bm\Theta_2 \circ \bm{R}_d +\bm{E}_2+\bm\Delta\bm{\Theta}_2\circ\bm\nabla \bm{R}_d -\bm{D}\rangle \\
& +\mu / 2\left\|\bm\Theta_2 \circ \bm{R}_d+\bm{E}_2+\bm\Delta \bm{\Theta}_2\circ\bm\nabla \bm{R}_d -\bm{D}\right\|_F^2 \\
& + \left\langle \bm{Y}_3,\left[\begin{array}{c}
\bm{C} \\
\bm{D}
\end{array}\right]-\bm{A} \right\rangle +\mu / 2\left\|\left[\begin{array}{c}
\bm{C} \\
\bm{D}
\end{array}\right]-\bm{A} \right\|_F^2,
\end{split}
\end{equation}
where $\bm{Y}_1 \in \mathbb{R}^{2M \times 1}$, $\bm{Y}_2 \in \mathbb{R}^{2M \times 1}$, and $\bm{Y}_3 \in \mathbb{R}^{2M \times 2}$ are the Lagrange multipliers, respectively. The operation $\langle \bm{A,B}\rangle$ represents the inner product between two matrices $\bm{A}$ and $\bm{B}$, and $ \mu$ ( $ \mu \geq 0$ ) represents the penalty coefficient.

\textbf{Update the matrix $\bm{A}$:} when other variables are fixed, the subproblem \textit{w.r.t.} $\bm{A}$ is:
\begin{equation}\label{eq:derive-of-A}
\min \|\bm{A}\|_{rank=1}
+\frac{\mu}{2\lambda} \left \| \bm{A} -
\left[\begin{array}{l}
\bm{C} \\
\bm{D}
\end{array}\right] -\frac{\bm{Y}_3}{\mu} \right\|_F^2,
\end{equation}
which can be solved by the singular value threshold
method. Specially, letting $ \bm{U} \bm{\Sigma}\bm{V}^\top $ be the Singular Value Decomposition(SVD) form of
$ \left[\begin{array}{l}
\bm{C} \\
\bm{D}
\end{array}\right] +\frac{\bm{Y}_3}{\mu} $, the solution to Eq.~\eqref{eq:derive-of-A} is as follows:
\begin{equation}\label{eq:solution-of-A}
\bm{A}^*=\bm{U}\mathcal{S}_{\lambda/\mu}(\bm{\Sigma})\bm{V}^\top,
\end{equation}
where $ \mathcal{S}_t(\bm{X})=\operatorname{sign}(\bm{X}) \times \max \{|\bm{X}-t|, 0\} $ is the shrinkage operator.

\textbf{ Update the matrix $\bm{C}$:} with other variables fixed, the subproblem \textit{w.r.t.} $\bm{M}$ is:
\begin{equation}\label{eq:derive-of-M}
\begin{split}
\bm{C}^*=\arg \min_{\bm{C}}  \frac{1}{2} \big \| & \bm{C} -
\frac{1}{4}(\bm\Theta_1 \circ \bm{P}+\bm{E}_1+\bm\Delta \bm\Theta_1\circ\bm\nabla \bm{P} \\
& +\bm{Y}_1 / \mu+\bm{A}_1-\bm{Y}_3^1/\mu) \big\|_F^2,
 \end{split}
\end{equation}
where $\bm{A}=[\bm{A}_1 : \bm{A}_2]^\top$,and $\bm{Y}_3=[\bm{Y}_3^1 : \bm{Y}_3^2 ]^\top$. Eq.~\eqref{eq:derive-of-M} is solved in a closed form solution as follows:
\begin{equation}\label{eq:solution-of-M}
  \bm{C}^*=  \frac{1}{4}\big(\bm\Theta_1 \circ \bm{P}+\bm{E}_1+\bm\Delta \bm\Theta_1\circ\bm\nabla \bm{P}
  +\bm{Y}_1 / \mu+\bm{A}_1-\bm{Y}_3^1/\mu\big).
\end{equation}

\textbf{ Update the matrix $\bm{D}$:} with other variables fixed, the subproblem \textit{w.r.t.}  $\bm{D}$ is:
\begin{equation}\label{eq:derive-of-D}
\begin{split}
\bm{D}^* = \arg \min_{\bm{D}} \frac{1}{2} \big \| &\bm{D} -
\frac{1}{4}(\bm\Theta_2\circ  \bm{R}_d+\bm{E}_2+\bm\Delta \bm\Theta_2\circ\bm\nabla \bm{R}_d \\
& +\bm{Y}_2 / \mu+\bm{A}_2-\bm{Y}_3^2 / \mu) \big \|_F^2.
\end{split}
\end{equation}
Eq.~\eqref{eq:derive-of-D} is solved in a closed form solution as follows:
\begin{equation}\label{eq:solution-of-D}
  \bm{D}^*=  \frac{1}{4}\big(\bm\Theta_2 \circ \bm{R}_d+\bm{E}_2+ \bm\Delta \bm\Theta_2\circ\bm\nabla \bm{R}_d  +\bm{Y}_2 / \mu+A_2-\bm{Y}_3^2 / \mu \big)
\end{equation}

\textbf{ Update the matrix $\bm{E}_1$:} with other variables fixed, the sub-problem \textit{w.r.t.} $\bm{E}_1$ is simplified as follows:

\begin{equation}\label{eq:derive-of-E1}
\begin{split}
\bm{E}^*_1=&\min_{\bm{E}_1} \|\bm{E}_1\|_1
 +\frac{\bm\mu}{2}  \| \bm{E}_1
- (\bm{C}-\\
&\bm\Theta_1 \circ \bm{P} - \bm\Delta \bm\Theta_1\circ\bm\nabla\bm{P}-\bm{Y}_1 / \mu)\|_F^2,
\end{split}
\end{equation}
which can be solved in a closed form solution,
\begin{equation}\label{eq:solution-of-E1}
 \bm{E}_1=S_{1 / \mu}\left(\bm{C}-\bm\Theta_1 \circ \bm{P}-\bm\Delta \bm\Theta_1\circ\bm\nabla\bm{P}-\bm{Y}_1 / \mu\right).
\end{equation}

\textbf {Update the matrix $\bm{E}_2$:} with other variables fixed, the sub-problem \textit{w.r.t.} $\bm{E}_2$ is as follows:
\begin{equation}\label{eq:derive-of-E2}
\min_{\bm{E}_2} \frac{\mu}{2} \left \| \bm{E}_2- \underbrace{(\bm{D}-\bm\Theta_2\circ   \bm{R}_d-\bm\Delta \bm\Theta_2\circ\bm\nabla\bm{R}_d  -\bm{Y}_2 / \mu)}_{\bm{E}} \right\|_F^2.
\end{equation}
%In the updating process, the variations of each element are unique, causing $\bm{E}_2$ to lose its translational characteristic.
The translational property makes $\bm{E}_2$ have the same column values. Therefore, $\bm{E}_2$ is solved as follows:
\begin{equation}\label{eq:solution-of-E2}
\bm{E}^*_2=\text{repmat}\left([\text{mean}(\bm{E})],\bm{C} \right),
\end{equation}
where the function $\text{repmat}(\bm{A},\bm{C})$ returns an array containing $C$ copies of $\bm{A}$, and the function $\text{mean}(\bm{A})$ returns an array containing the mean of the matrix $\bm{A}$ along the dimension of the row.

\textbf {Update the matrix $\bm{\Theta_1}$:} with other variables fixed, the derivative of Eq.~\eqref{eq:lagarian-multiplier-objective} with respect to $\bm{\Theta_1}$ should be 0, \textit{i.e.}, $\frac{\partial L}{\partial \bm{\bm\Delta\bm\Theta}_1} = 0$. The sub-problem \textit{w.r.t.} $\bm{\Theta_1}$:
\begin{equation}\label{eq:objective-of-theta_1}
\mu \bm\nabla \bm{P}^\top (\bm\Theta_1 \circ\bm{P}+\bm{E}_1+\bm\Delta \bm\Theta_1\circ\bm\nabla\bm{P}-\bm{C} +\frac{\bm{Y}_1}{\mu} ) = 0.
\end{equation}
The solution of Eq.~\eqref{eq:objective-of-theta_1} has a closed form as follows:
\begin{equation}\label{eq:solution-of-theta_1}
\bm\Delta\bm\Theta_1=(\bm\nabla\bm{P}^T\bm\nabla\bm{P})^{-1}\bm\nabla\bm{P}^\top[\bm{C}-(\bm\Theta_1 \circ\bm{P}+\bm{E}_1+\bm{Y}_1/\mu)].
\end{equation}

%Subsequently, the updated rotation angle is obtained as $\Theta_1 = \Theta_1 + \Delta \Theta_1$.

\textbf {Update the matrix $\bm{\Theta}_2$:} with other variables fixed, the derive of Eq.~\eqref{eq:lagarian-multiplier-objective} with respect to $\bm{\Theta}_2$ should be 0, i.e.,
$\frac{\partial L}{\partial \bm{\Delta\Theta}_2} = 0$. The sub-problem \textit{w.r.t.} $\bm{\Theta_2}$:
\begin{equation}\label{eq:objective-of-theta_2}
\mu \bm\nabla \bm{R}_d^\top (\bm\Theta_2 \circ\bm{R}_d+\bm{E}_2+\bm\Delta \bm\Theta_2\circ\bm\nabla\bm{R}_d-\bm{D} +\frac{\bm{Y}_2}{\mu}
) = 0
\end{equation}
The solution of Eq.~\eqref{eq:objective-of-theta_2} has a closed form as follows:
\begin{equation}\label{eq:solution-of-theta_2}
\bm\Delta\bm\Theta_{2}=(\bm\nabla\bm{R}_d^{\top}\bm\nabla\bm{R}_d)^{-1}\bm\nabla\bm{R}_d^\top[\bm{D}-(\bm\Theta_{2} \circ\bm{R}_d+\bm{E}_{2}+\bm{Y}_{2}/\mu)]
\end{equation}

\textbf {Update the Lagrange multiplier matrices $\bm{Y}_1$, $\bm{Y}_2$, and $\bm{Y}_3$.}
According the ADMM iteration in~\cite{BertsekasNonlinear}, the Lagrange multiplier matrices are update as follows:
\begin{eqnarray}
\bm{Y}_1& \leftarrow & \bm{Y}_1+\mu\left(\bm\Theta_1   \bm{P}+\bm{E}_1+ \bm \Delta \bm\Theta_1\bm\nabla\bm{P}-\bm{C}\right),\label{eq:objective-of-Y1}\\
\bm{Y}_2 & \leftarrow & \bm{Y}_2+\mu\left(\bm\Theta_2 \bm{R}_d+\bm{E}_2+ \bm\Delta \bm\Theta_2\bm\nabla\bm{R}_d-\bm{D}\right),\label{eq:objective-of-Y2}\\
\bm{Y}_3 & \leftarrow & \bm{Y}_3+\mu
\left(
\left[\begin{array}{l}
\bm{C} \\
\bm{D}
\end{array}\right]-\bm{A}\right)\label{eq:objective-of-Y3},
\end{eqnarray}
where $\mu\leftarrow \rho\cdot \mu $ ($\rho >1$) is a monotonically increasing positive sequence. Thus, we have transformed problem Eq.~\eqref{eq:lagarian-multiplier-objective} into a sequence of unconstrained
convex programs. We summarize the rank-1 alignment in Alg.~\ref{algorithmADMM}. The operations in each step of the algorithm are very simple. Note that SVD is still computationally efficient because not only the maximal rank of $\bm {A}$ is 2, but also the number of GPS points is relative small, \textit{e.g.},30.

\begin{small}
\begin{algorithm}[!t]
    \renewcommand{\algorithmicrequire}{\textbf{Input:}}
    \renewcommand{\algorithmicensure}{\textbf{Output:}}
    \caption{Optimize Eq.~\eqref{eq:lagarian-multiplier-objective} by ADMM}
    \label{algorithmADMM}
    \begin{algorithmic}[1]
        \Require
        The matrix $\bm{P} \in \mathbb{R}^{2M \times 1}$, the matrix $\bm{R_d} \in \mathbb{R}^{2M \times 1}$ for the parking spots on a map, and $\lambda > 0 $; % input
        \While {not converged}
           \State Update the matrix $\bm{A}$ by the singular value threshold method in Eq.~\eqref{eq:solution-of-A};
           \State
           Update the matrix $\bm{C}$ and $\bm{D}$ by Eq.~\eqref{eq:derive-of-M} and Eq.~\eqref{eq:derive-of-D}, respectively;
           \State
           Update the matrix $\bm{E}_1$ and $\bm{E}_2$ by Eq.~\eqref{eq:derive-of-E1} and Eq.~\eqref{eq:derive-of-E2}, respectively;
           \State
           Update the angle $\bm{\Theta_1}$ and $\bm{\Theta_2}$ by Eq.~\eqref{eq:solution-of-theta_1} and Eq.~\eqref{eq:solution-of-theta_2}, respectively;
           \State
           Update the Lagrange multiplier matrices $\bm{Y}_1$, $\bm{Y}_2$, and $\bm{Y}_3$ by Eq.~\eqref{eq:objective-of-Y1}-Eq.~\eqref{eq:objective-of-Y3};
           \State
           $\mu\leftarrow 1.3\mu$;
        \EndWhile
    \Ensure $\bm{E}_1$,$\bm{E}_2$,$\bm{A}$,$\bm\Delta\bm\Theta_1$, and $\bm\Delta\bm\Theta_2$. % output
 \end{algorithmic}
\end{algorithm}
\end{small}

\section{Experimental Results and Analysis}\label{sec:Experimental}

\subsection{Real Data}\label{sec:dataset}

\begin{figure}
    \centering
    \includegraphics[width=1\linewidth]{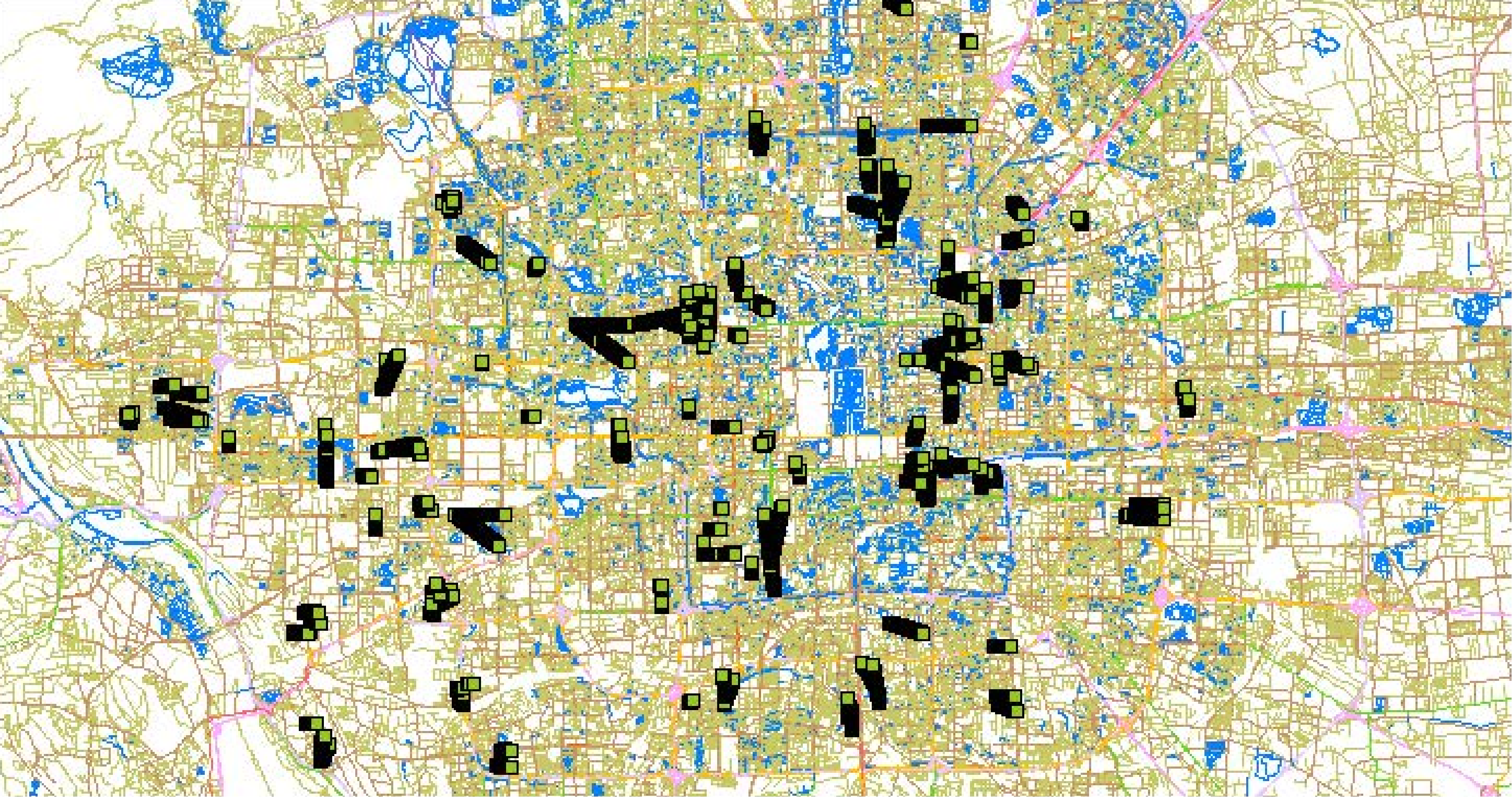}
    \caption{The parking spots in our dataset is projected onto a map. The boundary of the map is the longitude and latitude (116.3523566,39.86131312) to (116.6396339,39.96606271). The square blocks represent the parking spots we manually collected.}
    \label{fig:Parking Spot Dataset Map Projection}
\end{figure}

The dataset was sampled from approximately 20,000 roadside parking spots in Beijing, China. Specially, through rigorous quality control, we identified and excluded erroneous GPS points, ultimately selecting 81 road segments exhibiting significant positional distortion in parking spot localization. {\color{blue}The GPS points of parking spots are captured from a single-frequency equipment with one GPS point every 30 seconds. The distribution of the parking spots is shown in Fig.~\ref{fig:Parking Spot Dataset Map Projection}, which indicates that the GPS corruption is predominantly observed in urban areas.

The dataset consists of four components: 1) the manually collected error GPS points for parking spots; 2) the IDs of the road segments where the parking spots are located; 3) the GPS points of the road segments; 4) the attributes (\textit{e.g.}, intersection, road direction) of the road segments; and 5) the ground truth GPS points for road segments, collected by a dual-frequency GPS receiver with a 0.5 meter error tolerance.} The statistics of dataset are presented in Table~\ref{tbl:dataset}.

\begin{table}[!t]
\begin{center}
\caption{The statistics of parking spots dataset.}\label{tbl:dataset}
\scalebox{1.0}{
\begin{tabular}{|c|c|c|}
%\begin{tabular}{|m{2cm}<{\centering}|m{1.3cm}<{\centering}|m{1.3cm}<{\centering}|}
\hline
Type & \# Segments       & Average Parking Spots per Segment       \\ \hline\hline
Straight    & 18              & 39.09             \\ \cline{1-3}
Curve    & 63             & 58.58           \\ \hline
%Total number     & 81             & 54.37 \\ \hline
\end{tabular}
}
\end{center}
\end{table}

\subsection{Evaluation metrics} \label{sec:evaluation}

In our experiments, we use two evaluation criteria as follows:

\textbf{ Average point Coordinate Deviation} (ACD) is defined as follows:
\begin{equation}\label{eq:ACD}
ACD=\frac{\sum^{N}_{i=1} \sum^{M_i}_{j=1}\| \mathbf{p}_{i j}^{p d}- \mathbf{p}_{i j}^{g t }\|_2}{\sum^{N}_{i=1}M_i},
\end{equation}
{\color{blue}where $\mathbf{p}_{i j}^{g t}$ and $\mathbf{p}_{i j}^{p d}$ represent the ground truth and the rectified GPS points of the $j$-th parking spot on the $i$-th road segment, respectively, $M_i$ is the number of the parking spots on the $i$-th road segment, and $N$ is the number of road segments. The unit of ACD is meters ($m$). The smaller ACD is, a higher performance a method has. In this paper, we use ``$\uparrow$'' denotes that a higher value means a better performance; otherwise, ``$\downarrow$'' denotes that a lower value is a better performance.}

\textbf{Average Recall} (AR) is defined as follows:
\begin{equation}\label{eq:AR}
A R=\frac{\sum^{N}_{i=1} m_i^{p d} / m^{gt}_i}{N},
\end{equation}
where $ m_i^{pd}$ and $ m^{gt}_i $ represent the number of the accurately rectified parking spots and the number of parking spots on the $i$-th road segment, respectively. A GPS point is recognized as a successful rectification when the spatial distance $\|\mathbf{p}_{i}^{g t}-\mathbf{p}_{i}^{p d} \|_2$ is smaller than a threshold $\tau$. In this paper, $\tau=0.5 $ (\text{meter}) is used in our experiments.

\subsection{Methods in Comparison Study}\label{sec:Comparative Experiment}

%Compare the proposed approach with three state-of-the-art methods for topic detection on web:
We compare the proposed approach with the most conventional Point-to-Point (P2P) matching algorithms:
\begin{itemize}
\item The Euclidean Distance (ED) metric~\cite{danielsson1980euclidean}, which calculates the straight-line distance between point pairs, is widely employed in map-matching applications due to its computational efficiency and geometric interpretability. The ED between two GPS points $\mathbf{p}_i$ and $\mathbf{q}_i$ is as follows:
\begin{equation}\label{eq:DF-ED}
d_{ED}(\mathbf{p},\mathbf{q}) = \sqrt{\|\mathbf{p}_i - \mathbf{q}_i\|_2}.
\end{equation}
However, Eq.~\eqref{eq:DF-ED} is not the good choice to calibrate the non-rigid transformations, \textit{e.g.}, scale and rotation.

\item Chamfer Distance (CD)\cite{borgefors1988hierarchical} ~\cite{veeravasarapu2020proalignnet} is an approximate nearest-neighbor distance for two sets as follows:
\begin{equation}\label{eq:DF-CD}
d_{CD}(p,q) = \sum_{\mathbf{p}_i\in \mathcal{P}}\min_{ \mathbf{q}_j\in \mathcal{Q}} \|\mathbf{p}_i- \mathbf{q}_j\|_2 +
\sum_{\mathbf{q}_j\in \mathcal{Q}}\min_{ \mathbf{p}_i\in \mathcal{P}} \|\mathbf{p}_i- \mathbf{q}_j\|_2,
\end{equation}
where $ \mathcal{P}$ and $\mathcal{Q}$ are the two sets. The CD demonstrates robustness against noise and outliers; however, it fails to capture the intrinsic structural relationships embedded within points.

\item  Hungarian algorithm (HA)~\cite{mills2007dynamic} is an optimal assignment method by minimizing the total cost as follows:
\begin{equation}\label{eq:DF-HA}
\min \sum_{i=1}^N \sum_{j=1}^M \gamma_{ij} c(i, j),
\end{equation}
where  $\gamma_{ij}$ represents the weight of the task $i$ to the agent $j$, $c(i, j)$ is the cost of assigning the task  $i$  to the agent $j$, and $N$ and $M$ are the number of tasks and agents, respectively. The HA efficiently identifies the optimal pairing scheme in scenarios characterized by computationally prohibitive global optimization challenges

\item  Wasserstein distance (WD)~\cite{2000WD} measures the difference between two probability distributions as follows:
\begin{equation}\label{eq:DF-WD}
d_{WD}(\mathbf{p},\mathbf{q}) = \min_{\bm\gamma \in \bm\Gamma(\mathbf{p}_i\in \mathcal{P},\mathbf{q}_j\in \mathcal{Q})} \sum_{i,j} \bm\gamma_{ij} c(\mathbf{p}_i, \mathbf{q}_j),
\end{equation}
where  $\mathcal{P}$  and  $\mathcal{Q}$  are the two sets,  $\bm\Gamma(\mathbf{p}_i,\mathbf{q}_j)$  is the set of all possible coupling pairs, and $c(\mathbf{p}_i,\mathbf{q}_j)$ is the cost function. The WD is particularly well-suited for scenarios requiring quantitative assessment of distributional disparities.

\end{itemize}

\subsection{Qualitative Comparisons with The SOTA Methods}

\begin{table*}[t!]
\centering
\caption{Comparisons between the SOTA methods and our method. The best accuracy is in bold and the second best is underlined.}
\label{tab:matching performance}

\begin{tabular}{|c|c|c|c|c||c|c|}
\hline \multirow{2}*{Method}& \multicolumn{2}{c|}{Straight} & \multicolumn{2}{c||}{Curve} & \multicolumn{2}{c|}{All types} \\
\cline{2-7}
 & ACD $\downarrow$ & AR (\%) $\uparrow$ & ACD $\downarrow$ & AR (\%) $\downarrow$& ACD $\downarrow$ & AR (\%) $\uparrow$\\
\hline\hline
ED & 8.01 & 97.7  & \underline{13.23} & \underline{95.6 } & \underline{9.12} & 97.3  \\
\hline
CD & 16.50 & 95.4  & 62.21 & 86.3  & 26.40 & 94.0  \\
\hline
HA & 15.71 & 95.0  & 50.70 & 84.0  & 24.26 & 93.2  \\
\hline
WD & \underline{6.30} & \underline{97.7}  & 15.42 & 94.9  & 9.63 & \underline{97.3 }\\
\hline
Our method & $\bm{3.99}$ & $\bm{98.9 }$ & $\bm{12.62}$ & $\bm{95.8 }$ & $\bm{6.71}$ & $\bm{98.4 }$ \\
\hline
\end{tabular}
\end{table*}

Table~\ref{tab:matching performance} showed that  our model significantly outperforms the State-of-the-Art (SOTA) methods, \textit{i.e.}, ED, CD, HA, and WD, on the released dataset. {\color{blue} Our method consistently achieved the best results for all road segment types, with the lowest ACD value and  the highest AR one. It surpasses the second best method (\textit{i.e.}, ED) by $2.41$. The results suggest that our method successfully handles the different types of GPS point corruptions. Specifically, the experimental results demonstrate that our proposed method achieves significant performance improvements, reducing the ACD to 3.99 for straight road segments and 12.62 for curved segments. The comparisons demonstrate that our method outperforms WD and ED by a noticeable margin. The improvement is particularly pronounced in curved road segments, where positional drift typically exhibits greater magnitude, thereby demonstrating the superior robustness of our approach in handling complex road geometries.}

\begin{table*}[t!]
\centering
\caption{Comparative analysis of SOTA methods versus our proposed approach under random noise injection into GPS points, evaluated by the ACD and AR metrics.}
\label{tab:robustness performance}
\begin{tabular}{|c|c|c|c|c||c|c|}
\hline
\multirow{2}*{Method} & \multicolumn{2}{c|}{Straight} & \multicolumn{2}{c||}{Curve} & \multicolumn{2}{c|}{All types} \\ \cline{2-7}
 & ACD $\downarrow$ & AR (\%) $\uparrow$ & ACD  $\downarrow$ & AR (\%) $\uparrow$ & ACD $\downarrow$ & AR (\%)$\uparrow$ \\
\hline\hline
ED & 8.37 & \underline {96.6} & \underline{14.05} & \underline{94.3} & \underline{9.60} & \underline{96.3} \\
\hline
CD & 17.6 & 95.1 & 80.21 & 83.3 & 30.90 & 93.3 \\
\hline
HA & 16.61 & 94.7 & 48.27 & 83.5 & 24.37 & 92.9 \\
\hline
WD & \underline {8.21} &\underline {96.6} & 16.30 & 93.6 & 9.95 & 96.1 \\
\hline
Our method & $\bm{5.75}$ & $\bm{98.7}$ & $\bm{13.35}$ & $\bm{95.6}$ & $\bm{8.96}$ & $\bm{97.9}$\\
\hline
\end{tabular}
\end{table*}

\subsection{Robustness of Our method}\label{sec:Robustness}

To verify the robustness of our method, we impose random noise from the uniform distribution, \textit{i.e.}, $U\sim [0,20]$ (meters), to the manually collected GPS points of parking spots.
Table~\ref{tab:robustness performance} verified that the robustness of our method was slightly influenced by random noise. Compared with the scenario without the artificial noise, the ACD of our method increased from 6.71 to 8.96; while, the ACD of the second-best method (\textit{i.e.}, ED) increased from 9.12 to 9.60. Besides, our method consistently achieved the smallest ACD among SOTA methods. In terms of AR, our method still exhibited excellent robustness against artificial noise. For instance, the performance of our method decreased by 0.5\%, while the performances of ED, CD, HA, and WD decreased by 1\%, 0.7\%, 0.3\%, and 1.2\%, respectively. Although HA did not exhibit a significant performance decline, HA was still incomparable to our method (\textit{i.e.}, 92.9\% vs. 97.9\%).

To comprehensively assess performance, we integrate both robustness and accuracy into a unified evaluation metric as follows:
\begin{equation}\label{eqt:robust_indx}
 R(A)=\left\{
 \begin{array}{ll}
      \frac{|r^{n}(A)-r(A)|}{r(A)} & \text{if} \   \ A\  \ \text{is} \  \ \uparrow,\\
       \sqrt{|r^{n}(A)-r(A)|}\cdot r(A) &  \text{otherwise}\  \ A \  \ \text{is} \  \ \downarrow,
 \end{array}
 \right.
\end{equation}
where $r^{n}(A)$ and $r(A)$ denote the model's performance under metric $A$ with and without artificial noise injection into GPS points, respectively. Eq.~\eqref{eqt:robust_indx} utilizes the square function to non-linearly to scale the difference $|r^{n}(A)-r(A)|$.   Consequently, Eq.~\eqref{eqt:robust_indx} measures the robustness of a model in a normalized approach. The robustness of the model increases as the value of $R$ in Eq.~\eqref{eqt:robust_indx} decreases.

\begin{table}[h!]
    \centering
    \scriptsize
    \caption{Comparative analysis of SOTA methods versus our proposed approach under noise injection conditions, evaluated by the metric $R(A)$ in Eq.~\eqref{eqt:robust_indx}.}
    \label{tab:robustness Metric R}
    \begin{tabular}{|c|c|c|c|c|c|c|}
    \hline
    \multicolumn{2}{|c|}{Metric} & ED & CD & HA & WD & Our method  \\ \hline\hline
    \multirow{3}{*}{R(ACD)$\downarrow$} & Straight  & $\bm{4.81}$ & 17.31 & 14.90 & 8.71 & \underline{$\underline{5.29}$}  \\ \cline{2-7}
                               & Curve & $\underline{11.98}$ & 263.93 & 67.18 & 14.47 & $\bm{10.78}$  \\ \cline{2-7}
                               & All types & $\underline{6.32}$  & 56.00 & 24.51 & $\bm{5.45} $ &  10.07  \\ \hline\hline
    \multirow{3}{*}{R(AR)$\downarrow$} & Straight  & 1.13 & $\underline{0.31}$ & 0.32 & 1.13 & $\bm{0.20}$  \\ \cline{2-7}
                               & Curve & 1.36 & 3.48 & $\underline{0.60}$ & 1.37 & $\bm{0.21} $ \\ \cline{2-7}
                               & All types & 1.03 & 0.74 & $\bm{0.32} $  & 1.23 & \underline{$\underline{0.51}$}  \\ \hline
    \end{tabular}
\end{table}

Table~\ref{tab:robustness Metric R} illustrated the robustness of our method in terms of $R(A) $ in Eq.~\eqref{eqt:robust_indx}. Under various road segments (\textit{i.e.}, straight, and curve) and metrics (\textit{i.e.}, ACD and AR), our method consistently ranked first or second among SOTA approaches. For instance, on curved road segments, it demonstrated superior performance over the second-best methods (ED and HA) by margins of $1.20$ and $0.39$, respectively, while significantly outperforming all other baselines.

In summary, the results in Table~\ref{tab:robustness performance} and Table~\ref{tab:robustness Metric R} confirm that our method maintains superior performance in noisy environments, with minimal accuracy loss compared to baseline approaches.

\subsection{Sensitivity Analysis}

In this section, we examine the impact of the hyper-parameters, \textit{i.e.}, $\lambda$ in Eq.~\eqref{eqt:rank-1-align}, on the performance of a model.
We conducted ablation studies by assigning $\lambda$ with $1, 10, 100, 1000$, and $10000$, respectively.

\begin{table}[!h]
\caption{Effect of the different $\lambda$ on the performances of our method.}
\label{tab:weight coefficients}
\centering
\begin{tabular}{|cc|ccccc|}
\hline
\multicolumn{2}{|c|}{\multirow{2}{*}{Our method}  }    & \multicolumn{5}{c|}{$\lambda$}                                            \\ \cline{3-7}
\multicolumn{2}{|c|}{}                                    & \multicolumn{1}{l|}{1}     & \multicolumn{1}{l|}{10}   & \multicolumn{1}{l|}{\textbf{100}}   & \multicolumn{1}{l|}{1000}  & 10000 \\ \hline\hline
\multicolumn{1}{|c|}{\multirow{3}{*}{ACD $\downarrow$}}    & Straight  & \multicolumn{1}{l|}{16.34} & \multicolumn{1}{l|}{6.41} & \multicolumn{1}{l|}{\textbf{3.99}}  & \multicolumn{1}{l|}{5.15}  & 4.99  \\ \cline{2-7}
\multicolumn{1}{|c|}{}                        & Curve     & \multicolumn{1}{l|}{29.68} & \multicolumn{1}{l|}{13.6} & \multicolumn{1}{l|}{\textbf{12.6}} & \multicolumn{1}{l|}{13.75} & 20.12 \\ \cline{2-7}
\multicolumn{1}{|c|}{}                        & All types & \multicolumn{1}{l|}{18.37} & \multicolumn{1}{l|}{7.83} & \multicolumn{1}{l|}{\textbf{6.71}}  & \multicolumn{1}{l|}{6.84}  & 7.95  \\ \hline
\multicolumn{1}{|c|}{\multirow{3}{*}{AR(\%)$\uparrow$}} & Straight  & \multicolumn{1}{l|}{94.3}  & \multicolumn{1}{l|}{98.4} & \multicolumn{1}{l|}{\textbf{98.9}}  & \multicolumn{1}{l|}{98.8}  & 98.9  \\ \cline{2-7}
\multicolumn{1}{|c|}{}                        & Curve     & \multicolumn{1}{l|}{91.7}  & \multicolumn{1}{l|}{95.4} & \multicolumn{1}{l|}{\textbf{95.8}}  & \multicolumn{1}{l|}{95.4}  & 93.9  \\ \cline{2-7}
\multicolumn{1}{|c|}{}                        & All types & \multicolumn{1}{l|}{93.9}  & \multicolumn{1}{l|}{98.0}   & \multicolumn{1}{l|}{\textbf{98.4}}  & \multicolumn{1}{l|}{98.3}  & 98.2  \\ \hline
\end{tabular}
\end{table}

Table~\ref{tab:weight coefficients} indicated that the best choice of $ \lambda$ is 100. Concretely, the ACD and the AR of our method for all types road segments were $6.71$ and $98.4\%$, respectively. Importantly, we have the other observation: the performance of our method was significantly degraded when $ \lambda $ is equal to 1. Specially, at $ \lambda = 1$, the ACD and the AR for all types road segments were $18.37$ and $93.79\%$, respectively. The comparison results between $\lambda=100$ and $\lambda=1$ underscored the important role of the low-rank constraint to the RAA problem in our model. Furthermore, our method demonstrates robust performance across a wide range of
$\lambda$ values (\textit{i.e.}, from 10 to 1000), indicating  its hyperparameter insensitivity and ease of tuning.

%xxxx黑色点的label 匹配上的候选停车点
\begin{figure}[t!]
\centering
\subfloat[CD]{\includegraphics[width=1.6in]{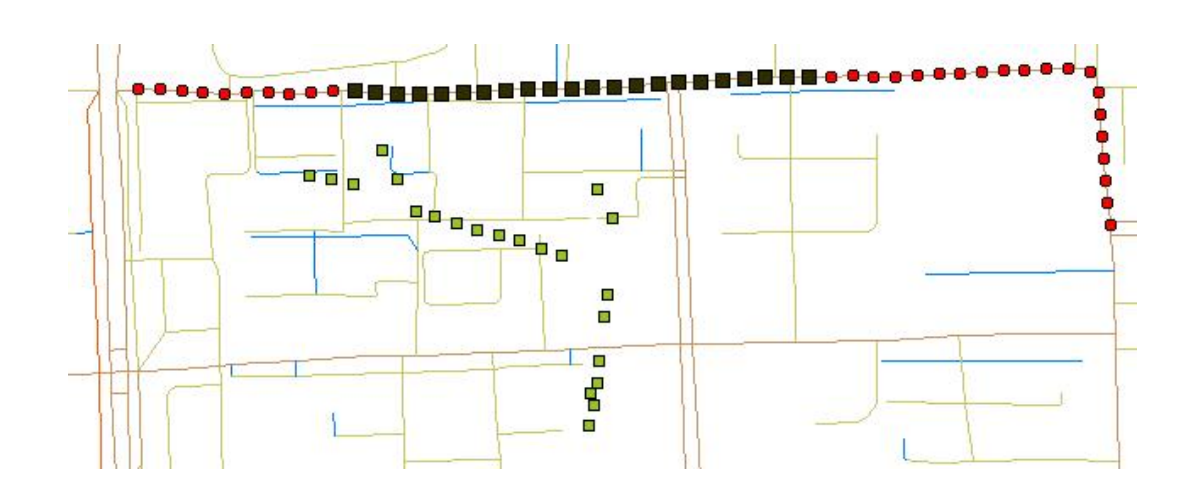}
}
\subfloat[ED]{\includegraphics[width=1.6in]{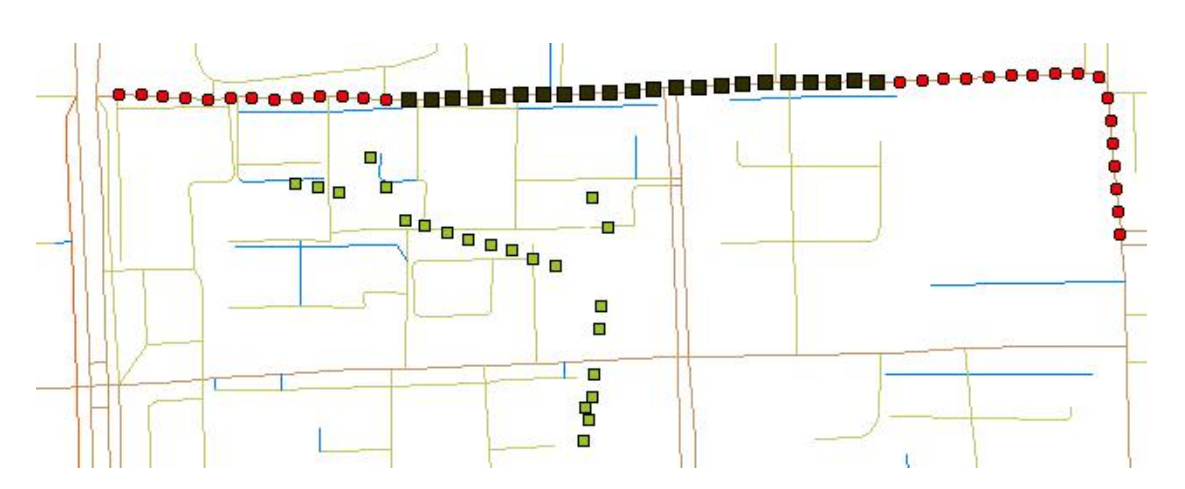}
}\\
\subfloat[HA]{\includegraphics[width=1.6in]{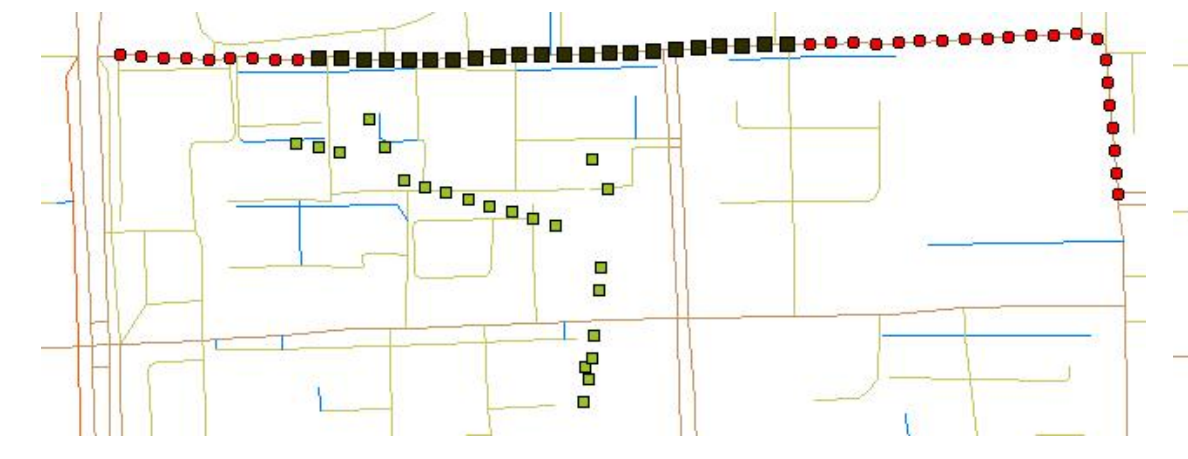}
}
\subfloat[WD]{\includegraphics[width=1.67in]{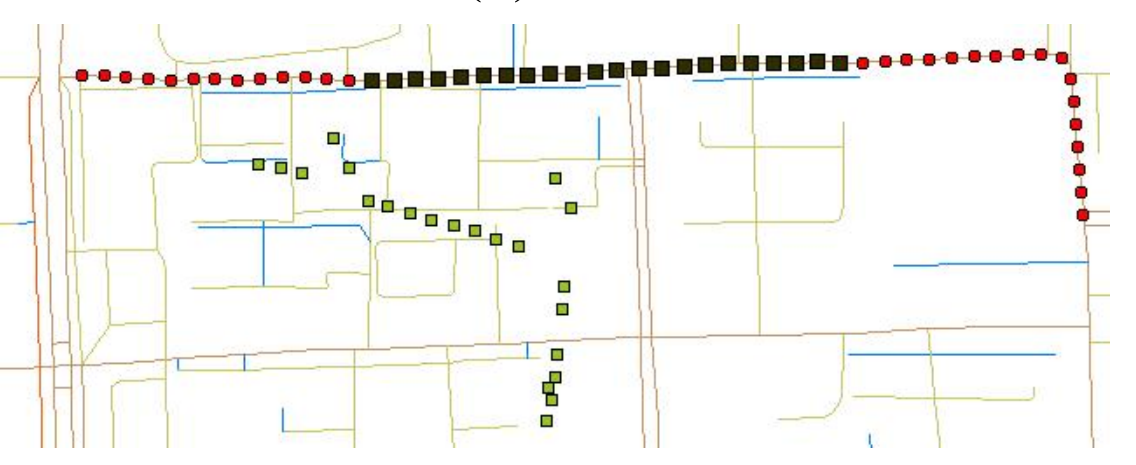}
}\\
\subfloat[Our method]{\includegraphics[width=1.67in]{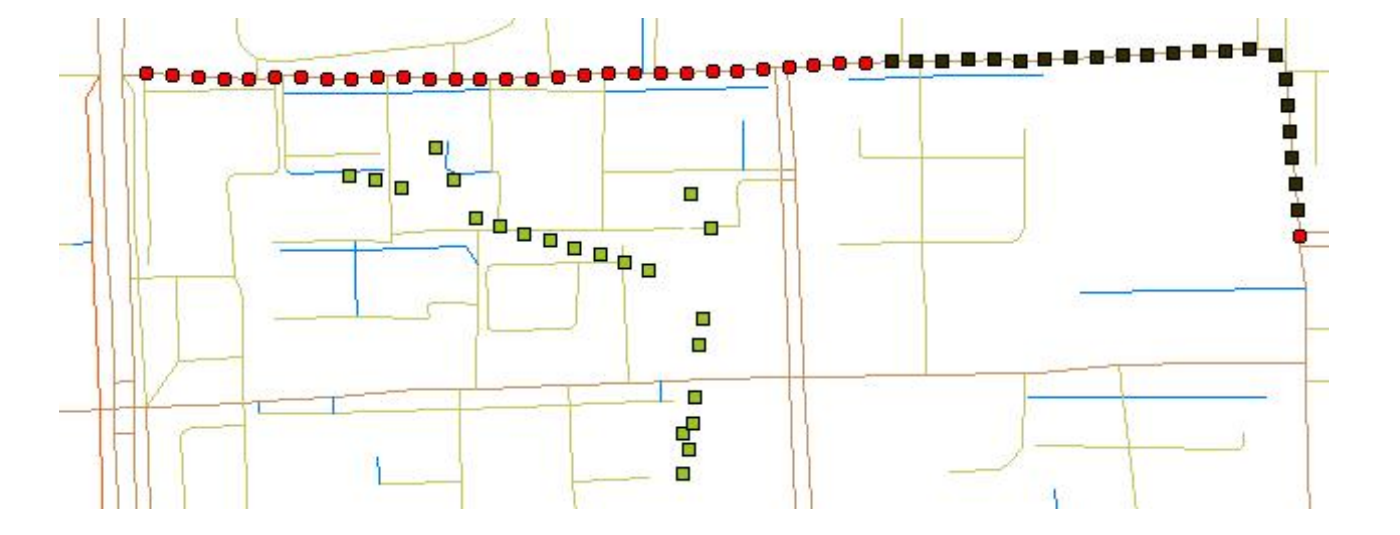}}
\subfloat[GT]{\includegraphics[width=1.6in]{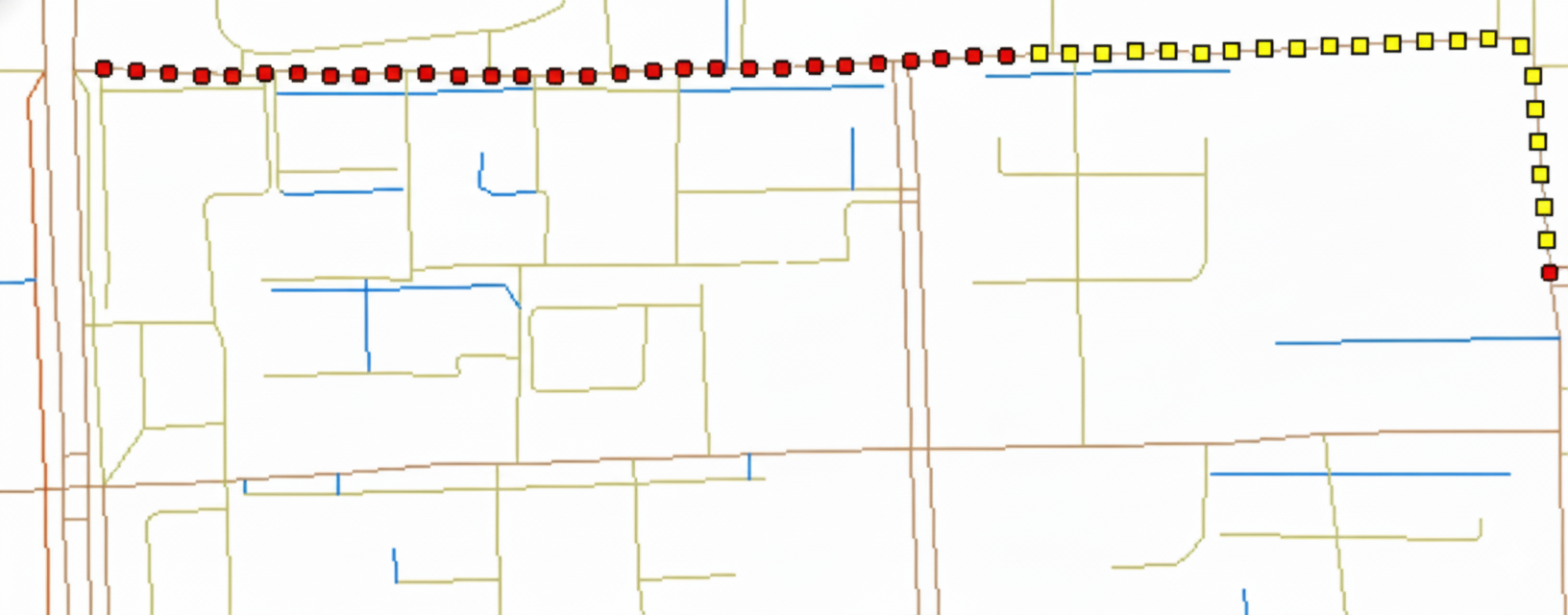}
}\\
\subfloat{\includegraphics[width=2.5in]{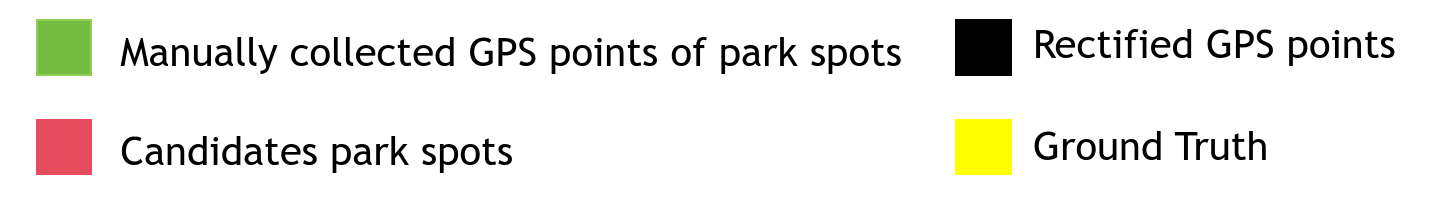}}
\caption{The manually collected GPS points of parking spots (in green) were rectified and aligned to the parking spots (in red) on a road segment, and the black points are the rectified and aligned GPS points for parking spots (best viewed in color).}
\label{fig:Visualization}
\end{figure}

\subsection{Visualization}\label{sec:Visual analytics }

To visually demonstrate the effectiveness of our method, we projected the corrected GPS coordinates onto a map. As illustrated in Figure~\ref{fig:Visualization}, our method exhibits significantly improved performance compared to existing state-of-the-art (SOTA) approaches. {\color{blue} Notably, the case shown in Fig.~\ref{fig:Visualization} exemplified a typical instance of positional errors, combining all three error types previously discussed. This representative scenario effectively demonstrated our method's capability in handling diverse real-world noise. Specifically, the manually collected GPS data in Fig.~\ref{fig:Visualization} contained sporadic corrupted points, conventional methods (CD, ED, HA, and WD in Fig.~\ref{fig:Visualization}(a)-(d)) failed to rectify these corruptions, just projecting them to the nearest valid points. In contrast, our approach (Fig.~\ref{fig:Visualization}(e)) successfully identified and corrected these outliers, accurately aligning GPS points with parking spots at curved road segments. The rectified results closely match the Ground Truth (GT) shown in Fig.~\ref{fig:Visualization}(f), visually confirming the method's superior error-correction capability.}

\begin{figure}[t!]
\centering
\subfloat[$U\sim\lbrack0,20\rbrack$]{\includegraphics[width=1.6in]{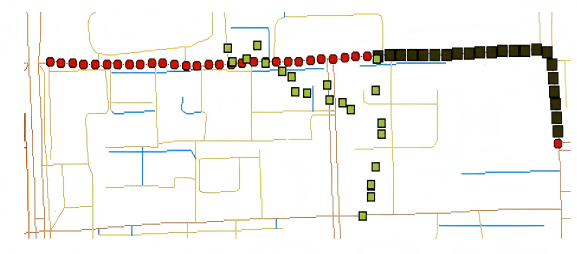}
}
\subfloat[$U\sim\lbrack0,50\rbrack$]{\includegraphics[width=1.6in]{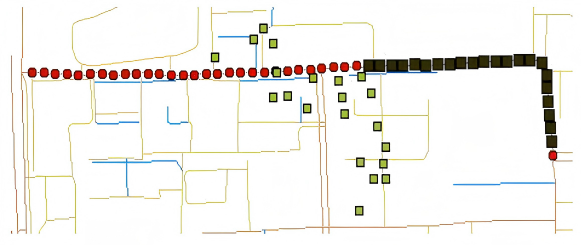}
}\\
\subfloat{\includegraphics[width=1.75in]{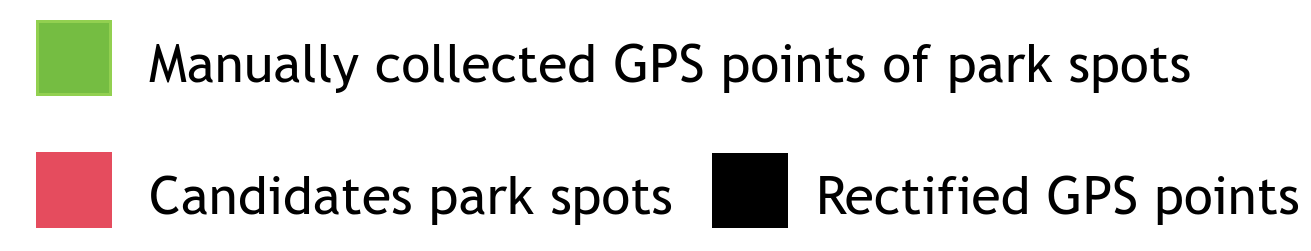}}
\caption{Our method demonstrates robust results when the different random noise are injected into the manually collected GPS points (best viewed in color).}
\label{fig:robustness test}
\end{figure}

To further examine the robustness of our model, we applied two different intense noise to the manually collected GPS points, \textit{i.e.}, one is the uniform distribution(\textit{i.e.}, $ U\sim [0,20]$), and the other is the uniform distribution(\textit{i.e.}, $U\sim [0,50]$). Fig.~\ref{fig:robustness test} showed that our method successfully rectified these corrupted GPS points into the corrected ones, despite the fact that different intensities of noise were injected. The comparison demonstrates that our method is commendably robust against random noise.

\section{CONCLUSION}\label{sec:conclusion}

This paper presents a novel low-rank constraint-based approach for robust alignment of GPS points to parking spots. Unlike conventional map-matching methods that rely on road segment topology or GPS point dynamics, our model utilizes static data sources exclusively, including manually collected GPS trajectories and road network attributes. To address diverse noise, the proposed framework jointly models translation/rotation transformations of corrupted GPS points and sparse outliers, employing low-rank matrix decomposition to enforce geometric consistency between rectified GPS points and parking spot coordinates. Extensive experiments on public datasets demonstrate superior noise resilience, achieving sub-meter-level matching accuracy with 98.4\% recall for all types of road segments. Statistical analysis confirms that our method is robust to both systematic and random GPS errors.

% \begin{figure}[t!]
%     \centering
%     \includegraphics[width=1.0\linewidth]{figures/robustness-test.png}
%     \caption{Our method demonstrates robust results when the different random noise are injected into the manually collected GPS points (best viewed in color).}
%     \label{fig:robustness test}
% \end{figure}

The demonstrated efficacy of our method in addressing the RAA problem motivates its extension to other applications within digital transportation platforms. Future work will focus on developing computational optimizations for the ADMM framework, such as the adaptive acceleration techniques proposed in~\cite{liu2021acceleration}, to further enhance algorithmic efficiency.

\bibliographystyle{IEEEtran}        % Include this if you use bibtex

\bibliography{dslref_vor1}

\end{document}